\documentclass[11pt]{article}
\usepackage{acl}
\usepackage{times}
\usepackage{latexsym}
\usepackage[T1]{fontenc}
\usepackage[utf8]{inputenc}
\usepackage{microtype}
\usepackage{graphicx}
\usepackage{amsmath,amssymb}
\usepackage{booktabs}
\usepackage{multirow}
\usepackage[table]{xcolor}

\usepackage{enumitem}
\usepackage{pifont}
\usepackage{float}
\usepackage{listings}
\usepackage{tabularx}
\usepackage{array}

\usepackage[most]{tcolorbox}
\tcbuselibrary{listings, breakable}

\definecolor{CAPSNavy}{HTML}{173A8A}
\definecolor{CAPSBlueBack}{HTML}{F8F9FD}
\definecolor{CAPSBlueFrame}{HTML}{D6DDEE}
\definecolor{CAPSBlueTitle}{HTML}{E9EEF8}
\definecolor{CAPSRowFill}{HTML}{E8EEF9}
\definecolor{CAPSGeneralFill}{HTML}{E8EEF9}
\definecolor{CAPSMedicalFill}{HTML}{EAF4EC}
\definecolor{CAPSLegalFill}{HTML}{EFEAF8}

\newtcblisting{promptbox}[2][]{%
  enhanced,
  breakable,
  colback=CAPSBlueBack,
  colframe=CAPSBlueFrame,
  colbacktitle=CAPSBlueTitle,
  coltitle=CAPSNavy,
  boxrule=0.45pt,
  arc=2pt,
  left=5pt,
  right=5pt,
  top=5pt,
  bottom=5pt,
  fonttitle=\bfseries\small,
  title={#2},
  listing only,
  before skip=6pt,
  after skip=8pt,
  listing options={
    basicstyle=\ttfamily\footnotesize,
    columns=fullflexible,
    keepspaces=true,
    breaklines=true,
    breakatwhitespace=true,
    breakautoindent=false,
    breakindent=0pt,
    showstringspaces=false,
    upquote=true,
    xleftmargin=0pt,
    framexleftmargin=0pt,
    aboveskip=0pt,
    belowskip=0pt
  },
  #1
}

\newtcolorbox{datacase}[1]{
  enhanced,
  before={\par\medskip\noindent\Needspace{12\baselineskip}},
  title={#1},
  width=\linewidth,
  colback=white,
  colframe=blue!35!black,
  colbacktitle=blue!8!white,
  coltitle=black,
  fonttitle=\bfseries\scriptsize,
  fontupper=\scriptsize,
  boxrule=0.5pt,
  arc=1.5pt,
  boxsep=2pt,
  left=4pt, right=4pt, top=2pt, bottom=2pt,
  toptitle=2pt, bottomtitle=2pt,
  before skip=6pt, after skip=6pt,
  parbox=false,
  segmentation style={solid, line width=0.4pt, blue!25!black},
}
\newcommand{\casefield}[1]{\noindent{\scriptsize\bfseries\color{blue!40!black}#1}\par}
\newcommand{\chosen}[1]{\noindent{\scriptsize\bfseries\color{green!50!black}[Chosen] #1}\par}
\newcommand{\rejected}[1]{\noindent{\scriptsize\bfseries\color{red!50!black}[Rejected] #1}\par}

\title{Criterion-Activated Preference Synthesis: \\Attribution-Preserving Preference Data from Explicit Standards}

\author{
\shortstack{
Zian Wang\thanks{Equal contribution.}\thanks{Work done during internship at Tencent.}\textsuperscript{1,2},
Pu Lu\footnotemark[1]\textsuperscript{1},
Caishuang Huang\textsuperscript{1},
Wenxi Wu\textsuperscript{1}
\\
Meng Zhou\textsuperscript{1},
Xiku Du\textsuperscript{1},
Yang Qiao\textsuperscript{1}
}
\\[0.35em]
\textsuperscript{1}Tencent \quad
\textsuperscript{2}The Hong Kong Polytechnic University
\\[-0.05em]
\small{\textbf{Correspondence:} \href{mailto:sunnyqiao@tencent.com}{sunnyqiao@tencent.com}}
}

\begin{document}

\raggedbottom
\maketitle

\begin{abstract}
Direct Preference Optimization (DPO) learns from chosen--rejected labels that can conflate accuracy, style, depth, and other differences, leaving the behavior behind a preference ambiguous. We propose \textbf{Criterion-Activated Preference Synthesis (CAPS)}, a criterion-first framework for constructing preference pairs around explicit quality standards. CAPS generates contexts and responses from a target criterion and scores each response with two views: a rubric-bound judge measures criterion compliance, while a blind judge measures holistic quality. The \textbf{Empirical Activation Test (EAT)} keeps contexts where satisfying the criterion improves blind quality; the \textbf{Attribution Ratio (AR)} then ranks pairs by how strongly their quality gap aligns with the criterion. Across general, medical, and Chinese legal criteria, CAPS shows strong data efficiency. With \textbf{20k} general synthetic pairs, it outperforms \textbf{35k} human-annotated HelpSteer3 pairs on IFEval, IFBench, AlpacaEval, and Creative Writing v3. CAPS also improves HealthBench with 2k medical pairs and improves Qwen3-8B/14B by +4.9 to +6.6 points on legal benchmarks using public statutes without human preference labels. We additionally construct the CAPS-Legal preference dataset.
\end{abstract}

\section{Introduction}
\begin{figure}[t]
\centering
\includegraphics[width=\columnwidth]{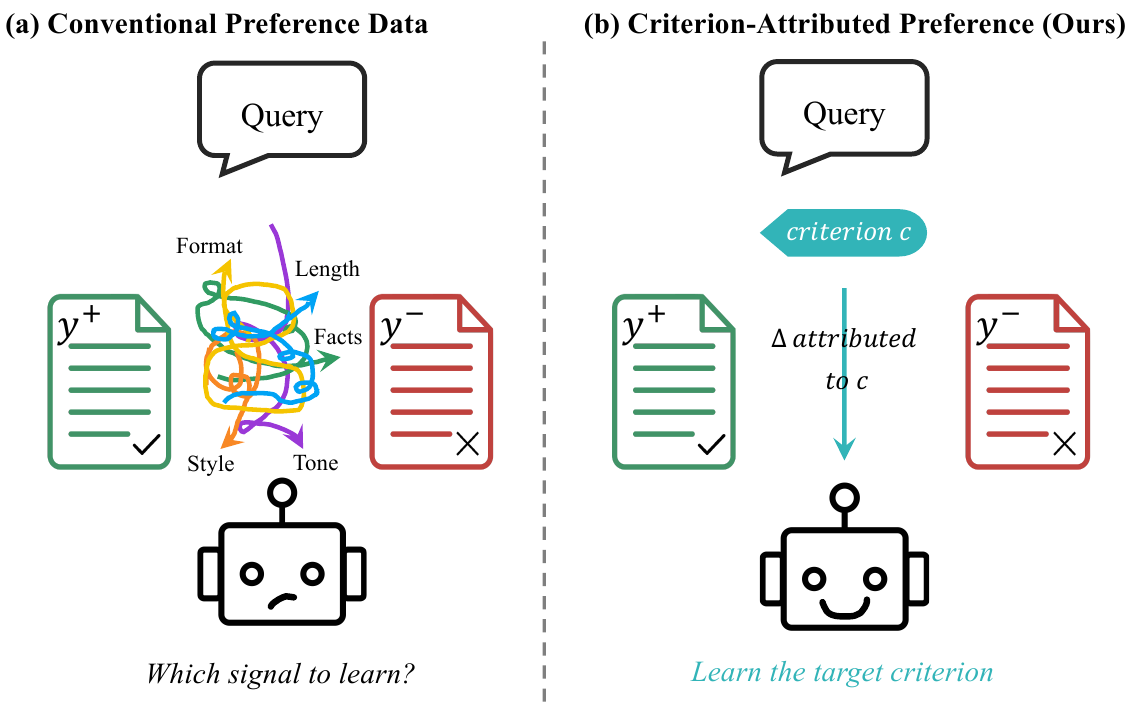}
\caption{Comparison between conventional preference data (left) and CAPS (right). Conventional methods produce pairs with confounded quality gaps across multiple dimensions---the model cannot determine which signal to learn. CAPS activates a specific criterion $c$ and verifies that the quality gap is primarily attributable to $c$ via dual-signal scoring.}
\label{fig:motivation}
\end{figure}

Direct Preference Optimization~\citep[DPO;][]{rafailov2024dpo} is simple to apply, but it learns only from a chosen--rejected label. Suppose a preferred answer cites primary sources, explains them more clearly, and is longer than the rejected answer. The label does not reveal whether citation quality, clarity, or length should be reinforced. We call this ambiguity about \emph{which behavior caused the preference} \emph{attribution ambiguity}.

Scaling such pairs does not necessarily resolve the ambiguity. Aggregating multi-model responses can instead amplify reward hacking~\citep{moreless2025}, while curated subsets often outperform larger unfiltered datasets~\citep{air2025,sweetspot2025,pvar2025,deita2024}. These findings motivate constructing pairs around an explicit target rather than assigning that target after collection.

We propose \textbf{Criterion-Activated Preference Synthesis (CAPS)}, a criterion-first pipeline. Given a criterion such as ``cite primary sources,'' CAPS first generates contexts where that criterion may matter and collects diverse responses. It then asks two separate questions. First, does satisfying the criterion improve overall response quality for this context? The \emph{Empirical Activation Test} (EAT) answers this at the context level and rejects contexts where the criterion is irrelevant or uninformative. Second, is a specific response pair's quality gap aligned with that criterion rather than obvious confounders? The \emph{Attribution Ratio} (AR) answers this at the pair level. Both checks use a rubric-bound score for criterion compliance and a blind score for holistic quality.

This design targets a narrower and directly testable objective: localizing preference gaps to explicit criteria and reducing confounding. It is motivated by prior analyses connecting preference-data rank deficiency to degenerate policies~\citep{rafailov2024dpo} and overoptimization~\citep{scalinglaws2024}, but we do not claim to measure representation rank or gradient alignment.

We instantiate CAPS with three criterion sources---HelpSteer3~\citep{wang2025helpsteer3preferenceopenhumanannotatedpreference} for general alignment, RaR Medicine~\citep{rar2025}, and Chinese legal statutes. Experiments show that 20k general CAPS pairs outperform 35k HelpSteer3 human pairs across general benchmarks. CAPS-Legal improves Qwen3-8B by +6.60 PLawBench~\citep{shi2026plawbenchrubricbasedbenchmarkevaluating} and +6.04 MSLR~\citep{yu2025benchmarkingmultisteplegalreasoning} over base, while the Infinity-Instruct legal subset degrades performance. Controlled scaling experiments further suggest moderate sample-efficiency gains in this setting.

Our contributions are threefold:
\begin{enumerate}[nosep,leftmargin=*]
    \item We introduce CAPS, a criterion-first framework that uses explicit standards to guide preference synthesis rather than only describing collected pairs post hoc.
    \item We propose complementary context- and pair-level checks. EAT tests whether criterion satisfaction corresponds to better blind quality in a context; AR ranks individual pairs by criterion alignment while penalizing large holistic gaps that may reflect confounders.
    \item We validate CAPS across general, medical, and Chinese legal criteria. In our general-domain experiments, 20k synthetic pairs outperform the 35k human-annotated HelpSteer3 baseline; CAPS also improves legal reasoning over the base models and the Infinity-Instruct legal subset baseline. We construct the CAPS-Legal preference dataset.
\end{enumerate}

\section{Methodology}
\label{sec:method}
\subsection{Overview}

\begin{figure*}[t]
\centering
\includegraphics[width=1\textwidth]{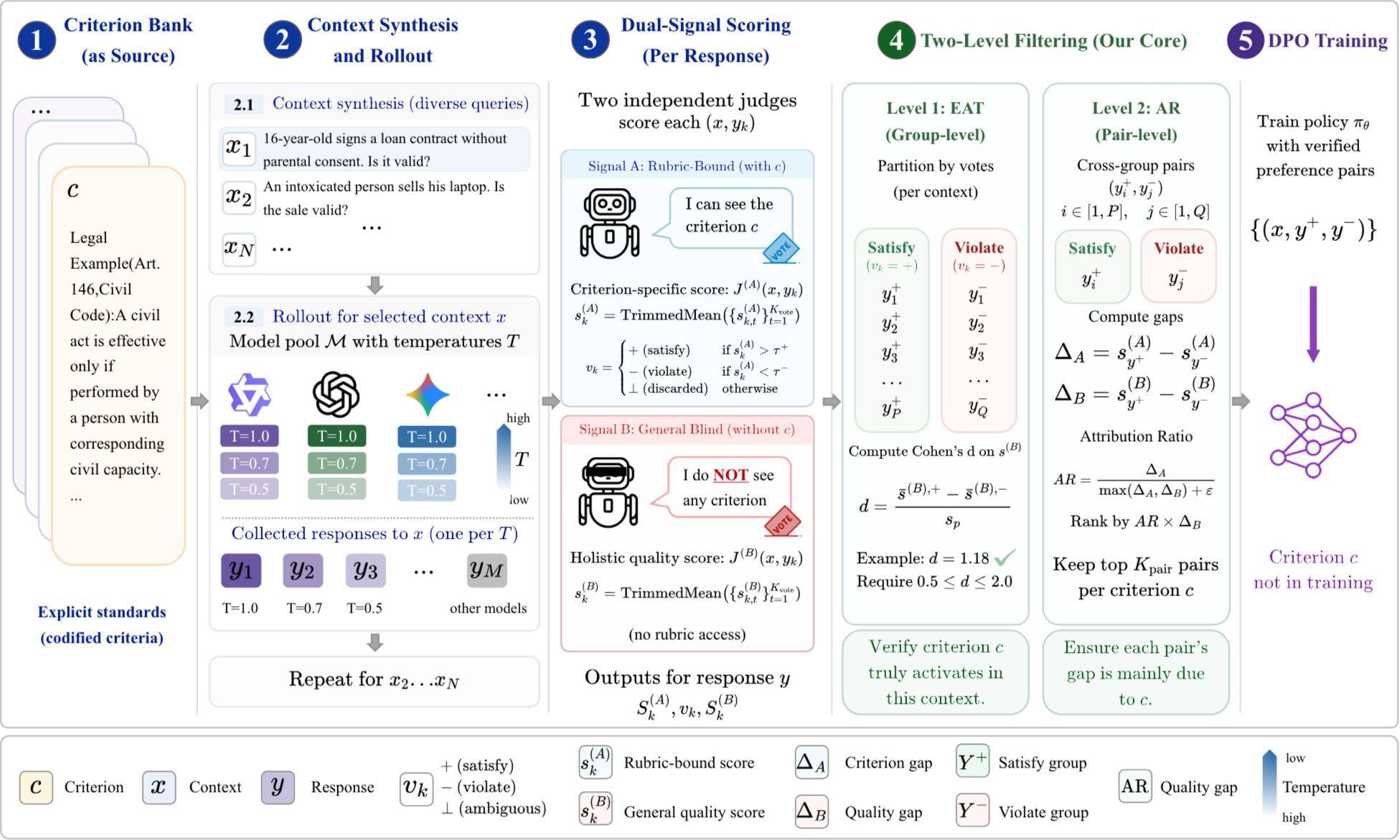}
\caption{Overview of the CAPS pipeline. Given criteria from a criterion bank, CAPS synthesizes attribution-verified preference pairs through context synthesis, multi-source rollout, and dual-signal scoring. EAT filters at the group level (criterion activation verification); AR filters at the pair level (attribution specificity ranking). The criterion $c$ does not enter the final DPO training---only the verified $(x, y^+, y^-)$ triplets are used.}
\label{fig:pipeline}
\end{figure*}

CAPS turns an explicit criterion into preference data in five steps: synthesize contexts where the criterion may apply, collect diverse response rollouts, score every response with criterion-aware and criterion-blind judges, filter uninformative contexts with EAT, and rank candidate pairs with AR. Intuitively, EAT asks whether the criterion matters for the context; AR asks whether it explains the direction of a particular pair. The checks are complementary: EAT is a group-level test and cannot detect confounding in one pair, whereas AR audits individual pairs but cannot determine whether a context is broadly informative. Figure~\ref{fig:pipeline} illustrates the pipeline, and Sections~\ref{sec:method:dualsignal}--\ref{sec:method:pair} formalize its components.

\subsection{Context Synthesis and Rollout}
\label{sec:method:context}

Given a criterion $c$ from a criterion bank $\mathcal{C}$, a generator model $G$ produces $N$ diverse contexts (user queries) where $c$ is potentially relevant, satisfying three requirements: (i)~\textbf{natural quality sensitivity}---$c$ compliance should naturally affect response quality; (ii)~\textbf{ecological validity}---contexts resemble realistic deployment queries; (iii)~\textbf{diversity}---variation in domain, complexity, and intent. Context synthesis makes \emph{no prior judgment} about whether $c$ will activate---this is deferred to EAT (\S\ref{sec:method:eat}), preventing confirmation bias.

For each context $x$, we collect responses from a diverse model pool $\mathcal{M}$ at multiple temperatures, yielding $M$ candidates (one per model-temperature combination). The multi-source design increases response diversity and prevents quality clustering.

\subsection{Dual-Signal Scoring}
\label{sec:method:dualsignal}

The core challenge is that no single signal can simultaneously verify (i)~\emph{whether} a response satisfies a given criterion and (ii)~\emph{whether} doing so improves perceived quality from a holistic standpoint. We therefore decompose preference judgment into two prompt-separated, functionally distinct signals: one explicitly conditioned on the criterion (Signal~A, rubric-bound), the other deliberately blind to it (Signal~B, general). Both signals follow the \emph{same} aggregation procedure---each judge is queried $K_{\text{vote}}$ times with stochastic decoding, the highest and lowest scores are trimmed, and the remainder is averaged. With $K_{\text{vote}}=3$, this aggregation reduces to the median and limits per-call judge variance.

\paragraph{Signal A (Rubric-Bound).} A judge $J^{(A)}$ scores each context-response pair $(x, y_k)$ on a 1--10 scale \emph{with full access to criterion $c$}, producing $K_{\text{vote}}$ stochastic samples $\{s_{k,t}^{(A)}\}_{t=1}^{K_{\text{vote}}}$. The aggregated score is
\begin{equation}
    s_k^{(A)} = \text{TrimmedMean}\big(\{s_{k,t}^{(A)}\}_{t=1}^{K_{\text{vote}}}\big),
    \label{eq:signal_a}
\end{equation}
which serves two purposes. First, $s_k^{(A)}$ feeds directly into the pair-level attribution score $\Delta_A$ (\S\ref{sec:method:pair}). Second, $s_k^{(A)}$ partitions responses into satisfy/violate groups via a double-sided threshold:
\begin{equation}
    v_k = \begin{cases}
        +\;\text{(satisfy)}  & \text{if } s_k^{(A)} > \tau^+ \\
        -\;\text{(violate)}  & \text{if } s_k^{(A)} < \tau^- \\
        \bot \;\text{(discarded)} & \text{otherwise}
    \end{cases}
\end{equation}
Responses falling in the ambiguous middle band ($v_k = \bot$) are excluded from subsequent stages, ensuring clean group separation for EAT. The rubric-bound paradigm follows PROMETHEUS~\citep{prometheus2023}, with trimmed-mean aggregation further stabilizing the decision boundary~\citep{Yin2018ByzantineRobustDL}.

\paragraph{Signal B (General Blind).} A judge $J^{(B)}$ scores the same response on a 1--10 scale \emph{without access to criterion $c$}, again with $K_{\text{vote}}$ stochastic samples and trimmed-mean aggregation:
\begin{align}
    s_k^{(B)} &= \text{TrimmedMean}\big(\{s_{k,t}^{(B)}\}_{t=1}^{K_{\text{vote}}}\big), \notag \\
    s_{k,t}^{(B)} &= J^{(B)}(x, y_k).
    \label{eq:signal_b}
\end{align}

Unlike Signal~A, Signal~B is \emph{not} thresholded and \emph{does not} produce a group label---it is consumed downstream as a continuous quality signal, used for both EAT's effect-size computation (\S\ref{sec:method:eat}) and AR's holistic gap $\Delta_B$ (\S\ref{sec:method:pair}). The judge evaluates quality holistically---covering fluency, factual accuracy, coherence, domain depth, and safety---a paradigm with established human-LLM agreement~\citep{openrs2025,zheng2024judging}. Because no rubric is provided, $s_k^{(B)}$ captures the \emph{total} perceived quality of the response, with any criterion-relevant or confounding effect on quality manifesting in this score---a property exploited by AR in \S\ref{sec:method:pair} to bound attribution. Issuing the signals as separate API calls with disjoint prompts and no shared context window establishes prompt-level information asymmetry, but does not make their model biases statistically independent.

\subsection{Empirical Activation Test (EAT)}
\label{sec:method:eat}

EAT decides: \emph{should this criterion-context pair enter the data pool at all?} Not every criterion produces a useful training signal in every context---a criterion about ``citing primary sources'' degenerates into noise when applied to a creative writing prompt. EAT asks: does satisfying $c$ measurably shift the perceived quality of responses to $x$? If yes, $(x, c)$ generates a learnable signal; if no, it is discarded before pair construction. This is a group-level decision made once per $(x, c)$ over all rollouts.

\paragraph{Group formation.} Using the labels $v_k$ from Signal~A, responses are partitioned into a satisfy group $Y^+ = \{y_k \mid v_k = +\}$ and a violate group $Y^- = \{y_k \mid v_k = -\}$, requiring $|Y^+| \geq 2$ and $|Y^-| \geq 2$ for valid variance estimation. Responses with ambiguous labels ($v_k = \bot$) are excluded. If a context fails to elicit both satisfying and violating responses (i.e., $|Y^+| < 2$ or $|Y^-| < 2$), it is deemed uninformative for criterion $c$ and is discarded---this prevents degenerate effect-size estimates from insufficient group contrast. If \emph{no} context passes EAT for criterion $c$, the criterion itself is excluded from the active bank.

\paragraph{Effect size computation.} For each group, we compute the mean and unbiased sample variance of Signal~B scores: $\bar{s}^{(B),+} = \frac{1}{n^+}\sum_{y_k \in Y^+} s^{(B)}_k$ (and analogously $\bar{s}^{(B),-}$), with sample variances $\sigma_+^2$ and $\sigma_-^2$ defined accordingly, where $n^+ = |Y^+|$ and $n^- = |Y^-|$. We then compute Cohen's $d$~\citep{cohen1977statistical} on these scores (crucially, \emph{not} Signal~A):
\begin{align}
    d &= \frac{\bar{s}^{(B),+} - \bar{s}^{(B),-}}{s_p}, \label{eq:cohens_d} \\
    s_p &= \sqrt{\frac{(n^+ - 1)\sigma_+^2 + (n^- - 1)\sigma_-^2}{n^+ + n^- - 2}} \notag
\end{align}
Computing $d$ on Signal~B avoids circularity: a large $d$ on Signal~A merely confirms the judge can read the rubric, whereas Signal~B measures whether criterion satisfaction translates to \emph{perceived quality improvement} from a blind evaluator---the property required for effective preference learning.

\paragraph{Double-sided threshold.} We require $d \in [\tau_{\min}, \tau_{\max}]$. Since $d$ is the ratio of group mean difference to pooled standard deviation $s_p$, extreme values arise from either component. When $d < \tau_{\min}$: either the criterion has no effect ($\bar{s}^{(B),+} \approx \bar{s}^{(B),-}$, criterion irrelevant), or the signal is drowned by within-group variance from multi-model heterogeneity or judge instability. When $d > \tau_{\max}$: either the quality gap is trivially large (e.g., on-topic vs.\ off-language responses, encouraging reward hacking~\citep{air2025,moreless2025}), or near-zero variance from few coarse-grained samples produces a statistical artifact ($s_p \to 0$, inflating $d$ despite a small absolute gap). See Appendix~\ref{app:eat} for detailed case analysis.

\subsection{Attribution Ratio (AR) and Pair Selection}
\label{sec:method:pair}

AR decides: \emph{is this specific pair's preference signal actually about the criterion?} Passing EAT confirms that $c$ activates \emph{on average} in context $x$, but says nothing about whether any \emph{particular} pair isolates the criterion. A randomly drawn pair from an EAT-passing context might differ predominantly in fluency or formatting---not in criterion compliance. AR provides a per-pair audit: by comparing the criterion-specific gap $\Delta_A$ against the holistic gap $\Delta_B$, it quantifies how much of the observable preference is attributable to $c$ versus to confounders. This is also the central mechanism by which CAPS mitigates multi-model reward hacking~\citep{moreless2025}: stylistic differences across model scales inflate $\Delta_B$ without proportionally increasing $\Delta_A$, yielding low AR scores that filter out confound-driven pairs (see Appendix~\ref{app:multimodel}).

For a candidate pair $(y^+, y^-)$ where $y^+ \in Y^+$ and $y^- \in Y^-$, we define:
\begin{equation}
    \Delta_A = s^{(A)}_{y^+} - s^{(A)}_{y^-}, \quad \Delta_B = s^{(B)}_{y^+} - s^{(B)}_{y^-}
\end{equation}
where $\Delta_A$ captures the criterion-specific score gap and $\Delta_B$ captures the holistic quality gap. $\Delta_A > 0$ is guaranteed by construction since $y^+ \in Y^+$ (satisfies $c$) and $y^- \in Y^-$ (violates $c$). $\Delta_B > 0$ is enforced via the directional consistency constraint below.

\paragraph{Attribution Ratio (AR).} We define:
\begin{equation}
    \text{AR} = \frac{\Delta_A}{\max(\Delta_A, \Delta_B) + \varepsilon} \in [0, 1]
    \label{eq:ar}
\end{equation}
where $\varepsilon = 10^{-8}$ prevents division by zero. The $\max$ operator normalizes the ratio, guaranteeing $\text{AR} \in [0, 1]$ regardless of whether $\Delta_A$ or $\Delta_B$ is larger---when $\Delta_A \geq \Delta_B$, $\text{AR} = \Delta_A/(\Delta_A + \varepsilon) \approx 1$ for $\Delta_A \gg \varepsilon$; when $\Delta_B > \Delta_A$, the denominator equals $\Delta_B + \varepsilon$ and AR decreases proportionally. The AR admits a natural interpretation:
\begin{itemize}[nosep,leftmargin=*]
    \item $\text{AR} \geq \tau_\text{AR}$ (retained): the criterion-specific gap is a substantial driver of the quality difference. Higher AR indicates stronger criterion dominance.
    \item $\text{AR} < \tau_\text{AR}$ (filtered): confounding factors dominate ($\Delta_B \gg \Delta_A$). Such pairs are excluded.
\end{itemize}
AR is used strictly as a ranking metric rather than a calibrated proportion, since LLM judge scores are fundamentally ordinal~\citep{prometheus2023}. Its empirical effectiveness is validated in \S\ref{sec:system_eval}, where high-AR pairs consistently outperform low-AR pairs at fixed data budget.

\paragraph{Pair quality constraints.} Beyond $\text{AR} \geq \tau_\text{AR}$, we impose four additional constraints grounded in empirical findings:
\begin{itemize}[nosep,leftmargin=*]
    \item \textbf{Directional consistency:} We strictly require $\Delta_A > 0$ and $\Delta_B > 0$---the chosen response must score higher on \emph{both} the criterion-specific and holistic dimensions. This eliminates paradoxical pairs where criterion satisfaction coincides with overall quality degradation.
    \item \textbf{Chosen quality:} $s^{(B)}_{y^+} \geq q_{\min}$, ensuring the chosen response has adequate overall quality rather than being merely ``less bad''~\citep{air2025}.
    \item \textbf{Margin bounds:} $\Delta_B \in [\Delta_{\min}, \Delta_{\max}]$, avoiding both noisy pairs (gap too small) and trivial pairs (gap too large)~\citep{sweetspot2025}.
    \item \textbf{Diversity cap:} each response appears in at most $K_{\text{rep}}$ selected pairs, preventing a small set of responses from dominating the data~\citep{rdpo2025}.
\end{itemize}

\paragraph{Ranking and selection.} Valid pairs are ranked by:
\begin{equation}
    \text{score}(y^+, y^-) = \text{AR} \times \Delta_B
    \label{eq:pair_score}
\end{equation}
This jointly rewards \emph{attribution purity} (high AR) and \emph{signal strength} (high $\Delta_B$). Neither alone suffices: $\text{AR} = 1$ with $\Delta_B \approx 0$ contributes negligible gradient, while high $\Delta_B$ with low AR teaches confound-driven preferences. We select the top $K_{\text{pair}}$ pairs per criterion-context combination $(x, c)$.

\paragraph{Complementarity.} EAT and AR address distinct failure modes: EAT-only filtering misses per-pair confounding (e.g., model identity driving quality gaps within an activated context~\citep{moreless2025}); AR-only filtering misses contexts where the criterion fails to activate entirely---without EAT, all contexts advance to pair selection regardless of whether criterion satisfaction actually correlates with quality, producing nominally high-AR pairs from judge noise in uninformative contexts. Together, EAT establishes that \emph{signal exists}; AR confirms that \emph{signal is correctly attributed}. We empirically validate this in \S\ref{sec:system_eval}.

\subsection{Training}
\label{sec:method:training}

We train the target policy $\pi_\theta$ on the selected pairs via standard DPO~\citep{rafailov2024dpo}; criterion $c$ is \emph{not} used during training. Holding the training objective fixed isolates the contribution of data construction from algorithmic novelty and makes CAPS compatible with variants such as IPO~\citep{azar2024general} and KTO~\citep{ethayarajh2024kto}.

\section{Experiments}
\label{sec:experiments}

\begin{table*}[!t]
\centering
\fontsize{7.2pt}{8.0pt}\selectfont
\setlength{\tabcolsep}{1.6pt}
\renewcommand{\arraystretch}{0.98}

\begin{minipage}[t]{0.515\textwidth}
\centering
\textbf{(a) General Benchmarks}
\vspace{2pt}

\scalebox{0.975}{%
\begin{tabular}{@{}llcccccc@{}}
\toprule
& & \multicolumn{4}{c}{\textbf{Instruction Following}} & \multicolumn{2}{c}{\textbf{Open-ended}} \\
\cmidrule(lr){3-6} \cmidrule(lr){7-8}
\multicolumn{1}{c}{\textbf{Method}} & \textbf{Data} & IFEval$_S$ & IFEval$_L$ & IFB$_S$ & IFB$_L$ & Alpaca & Creative \\
\midrule
\multicolumn{8}{@{}l}{\textit{Qwen3-8B}} \\
\quad Base        & --  & 82.26 & 85.95 & 25.67 & 30.00 & 57.00 & 59.00 \\
\quad HelpSteer3  & 35k & 84.47 & 87.43 & 32.33 & 36.67 & 66.51 & 58.96 \\
\quad CAPS        & \cellcolor{CAPSGeneralFill}20k
& \cellcolor{CAPSGeneralFill}\textbf{86.32}
& \cellcolor{CAPSGeneralFill}\textbf{89.46}
& \cellcolor{CAPSGeneralFill}\textbf{35.00}
& \cellcolor{CAPSGeneralFill}\textbf{38.33}
& \cellcolor{CAPSGeneralFill}\textbf{68.59}
& \cellcolor{CAPSGeneralFill}\textbf{74.53} \\

\midrule
\multicolumn{8}{@{}l}{\textit{Qwen3-14B}} \\
\quad Base        & --  & 85.40 & 88.54 & 27.67 & 30.00 & 63.23 & 70.42 \\
\quad HelpSteer3  & 35k & 85.03 & 88.17 & 34.67 & 40.00 & 71.23 & 68.13 \\
\quad CAPS        & \cellcolor{CAPSGeneralFill}20k
& \cellcolor{CAPSGeneralFill}\textbf{87.43}
& \cellcolor{CAPSGeneralFill}\textbf{90.02}
& \cellcolor{CAPSGeneralFill}\textbf{38.33}
& \cellcolor{CAPSGeneralFill}\textbf{42.67}
& \cellcolor{CAPSGeneralFill}\textbf{72.49}
& \cellcolor{CAPSGeneralFill}\textbf{79.66} \\

\midrule
\multicolumn{8}{@{}l}{\textit{Reference}} \\
\quad Qwen3-32B       & -- & 84.83 & 88.17 & 26.33 & 30.67 & 69.59 & 77.04 \\
\quad Qwen3-235B      & -- & 87.43 & 90.20 & 39.00 & 41.67 & 82.61 & 86.48 \\
\quad Gemma-4-31B     & -- & 91.87 & 94.09 & 49.00 & 52.67 & 77.26 & 86.00 \\
\bottomrule
\end{tabular}%
}
\end{minipage}
\hfill
\begin{minipage}[t]{0.205\textwidth}
\centering
\textbf{(b) Medical}
\vspace{2pt}

\scalebox{0.96}{%
\begin{tabular}{@{}lcc@{}}
\toprule
\multicolumn{1}{c}{\textbf{Method}} & \textbf{Data} & \textbf{Health} \\
\midrule
\multicolumn{3}{@{}l}{\textit{Qwen3-8B}} \\
\quad Base       & --  & 41.53 \\
\quad HelpSteer3 & 35k & 45.19 \\
\quad RaR-Med.   & 22k & 41.79 \\
\quad CAPS       & \cellcolor{CAPSMedicalFill}2k
& \cellcolor{CAPSMedicalFill}\textbf{45.30} \\

\midrule
\multicolumn{3}{@{}l}{\textit{Qwen3-14B}} \\
\quad Base       & --  & 45.34 \\
\quad HelpSteer3 & 35k & 47.59 \\
\quad RaR-Med.   & 22k & 44.07 \\
\quad CAPS       & \cellcolor{CAPSMedicalFill}2k
& \cellcolor{CAPSMedicalFill}\textbf{47.91} \\

\midrule
\multicolumn{3}{@{}l}{\textit{Reference}} \\
\quad Qwen3-32B   & -- & 44.36 \\
\quad Qwen3-235B  & -- & 51.26 \\
\quad Gemma-4-31B & -- & 48.93 \\
\bottomrule
\end{tabular}%
}
\end{minipage}
\hfill
\begin{minipage}[t]{0.26\textwidth}
\centering
\textbf{(c) Legal}
\vspace{2pt}

\scalebox{0.96}{%
\begin{tabular}{@{}lccc@{}}
\toprule
\multicolumn{1}{c}{\textbf{Method}} & \textbf{Data} & \textbf{PLaw} & \textbf{MSLR} \\
\midrule
\multicolumn{4}{@{}l}{\textit{Qwen3-8B}} \\
\quad Base       & --  & 41.29 & 55.58 \\
\quad HelpSteer3 & 35k & 44.94 & 45.03 \\
\quad Inf-Inst   & 10k & 26.39 & 31.43 \\
\quad CAPS       & \cellcolor{CAPSLegalFill}6k
& \cellcolor{CAPSLegalFill}\textbf{47.89}
& \cellcolor{CAPSLegalFill}\textbf{61.62} \\

\midrule
\multicolumn{4}{@{}l}{\textit{Qwen3-14B}} \\
\quad Base       & --  & 46.31 & 60.37 \\
\quad HelpSteer3 & 35k & 49.44 & 47.66 \\
\quad Inf-Inst   & 10k & 39.98 & 46.33 \\
\quad CAPS       & \cellcolor{CAPSLegalFill}6k
& \cellcolor{CAPSLegalFill}\textbf{51.25}
& \cellcolor{CAPSLegalFill}\textbf{65.77} \\

\midrule
\multicolumn{4}{@{}l}{\textit{Reference}} \\
\quad Qwen3-32B   & -- & 53.06 & 65.08 \\
\quad Qwen3-235B  & -- & 63.51 & 75.63 \\
\quad Gemma-4-31B & -- & 57.00 & 67.42 \\
\bottomrule
\end{tabular}%
}
\end{minipage}

\caption{Main results with separated general, medical, and legal evaluations. Panel (a) reports general alignment benchmarks, where CAPS uses the 20k general checkpoint. Panel (b) reports HealthBench, where CAPS uses the 2k medical checkpoint. Panel (c) reports legal benchmarks, where CAPS uses the 6k legal checkpoint. Qwen3-235B denotes Qwen3-235B-A22B-Instruct-2507. Shading indicates the corresponding CAPS data/checkpoint. \textbf{Bold} = best within each base-model block; $S$=strict, $L$=loose.}
\label{tab:main_results_three_panel_uniform}
\end{table*}

We conduct a comprehensive evaluation of CAPS across three domains: general-purpose alignment, healthcare, and practical legal tasks.

\subsection{Experimental Setup}

\paragraph{Datasets and criterion sources.} CAPS is instantiated on three criterion sources and synthesizes domain-specific preference pairs from each: (1)~\textbf{General}: 35k pair-level criteria extracted from HelpSteer3 by Auto-Rubric $\rightarrow$ 20k synthesized preference pairs; (2)~\textbf{Medical}: 22k pair-level criteria from RaR Medicine~\citep{rar2025} $\rightarrow$ 2k synthesized preference pairs\footnote{The lower medical conversion rate is mainly due to EAT filtering: RaR-Medicine-derived rubrics often target multi-step reasoning, where activation requires highly specific contexts.}; (3)~\textbf{Legal}: $\sim$2k Chinese statutory provisions $\rightarrow$ 6k synthesized preference pairs (corpus details in Appendix~\ref{app:caps_legal_profile}). General/Medical criteria are pair-level, while Legal criteria are statutory provisions that can instantiate multiple contexts. Representative preference pairs from each domain are shown in Appendix~\ref{app:data_examples}. We train one domain-specific DPO checkpoint per base model and evaluate it on the corresponding benchmarks.

\paragraph{External baselines.} We use external systems to contextualize downstream performance: Qwen3-8B/14B base models~\citep{yang2025qwen3technicalreport}; DPO on 35k human-annotated HelpSteer3 pairs for general benchmarks and HealthBench~\citep{arora2025healthbench}; SFT on RaR-Medicine~\citep{rar2025} for HealthBench; and SFT on 10k Infinity-Instruct legal examples~\citep{infinityinstruct2025} for PLawBench/MSLR (Appendix~\ref{app:inf_inst_legal}). Stronger zero-shot models---Qwen3-32B, Qwen3-235B-A22B-Instruct-2507, and Gemma-4-31B---provide additional scale references. Because these systems differ in source data or training objective, they are practical performance references rather than controlled mechanism comparisons.

\paragraph{Controlled comparisons.} Section~\ref{sec:system_eval} compares Random selection, either judge alone, and EAT/AR ablations using the same response rollout pool. These controls isolate the contribution of CAPS's verification and selection stages.

\paragraph{Metrics.} IFEval~\citep{zhou2023instruction} and IFBench~\citep{pyatkin2025generalizingverifiableinstructionfollowing}(strict/loose instruction-following), AlpacaEval~\citep{dubois2025lengthcontrolledalpacaevalsimpleway} LC, Creative Writing v3~\citep{paech2024eqbenchemotionalintelligencebenchmark}, HealthBench (clinical reasoning), PLawBench and MSLR (legal reasoning/retrieval).

\paragraph{Implementation.} We use DeepSeek-V3.2~\citep{deepseekai2025deepseekv32pushingfrontieropen} in non-thinking mode\footnote{DeepSeek API thinking-mode guide: \url{https://api-docs.deepseek.com/guides/thinking_mode}.} for context synthesis and both judge prompts. We generate 10 contexts per general/legal criterion and 2 per medical criterion, with 20 response rollouts per context. Defaults are Signal-A thresholds $(\tau^+,\tau^-)=(7,7)$, EAT interval $[0.5,2.0]$, and $\tau_{\mathrm{AR}}=0.6$. CAPS uses standard DPO; the Infinity-Instruct legal baseline uses SFT. All downstream results are from a single training run. Appendix~\ref{app:hyperparams} gives full training and synthesis settings, and Appendix~\ref{app:reproducibility} reports resources and costs.

\subsection{Main Results}
Table~\ref{tab:main_results_three_panel_uniform} summarizes results across the general, medical, and legal settings. CAPS improves the corresponding base models in all three domains while using compact synthesized preference sets.

\paragraph{General tasks.} CAPS consistently improves over base and outperforms the 35k HelpSteer3 DPO baseline with only 20k synthetic pairs. The most salient gain is Creative Writing, where CAPS-8B approaches Qwen3-32B (74.53 vs.\ 77.04) and CAPS-14B surpasses it (79.66 vs.\ 77.04). This pattern is consistent with CAPS reinforcing controllable aspects of open-ended generation---instruction realization, structure, and style---rather than only optimizing narrow instruction-following scores. As a result, CAPS improves Creative Writing without degrading other general benchmarks.

\paragraph{Medical domain.} CAPS improves HealthBench over base at both scales and roughly matches the much larger 35k HelpSteer3 DPO baseline using only 2k medical pairs (45.30 vs.\ 45.19 for 8B; 47.91 vs.\ 47.59 for 14B), which we treat as data-efficient parity. More notably, CAPS outperforms direct SFT on the 22k RaR-Medicine subset, which provides little benefit for 8B and degrades 14B, suggesting that verified preference conversion is a safer use of this source than direct instruction tuning. The modest gains likely reflect source--benchmark mismatch: RaR-Medicine focuses on clinical reasoning \citep{rar2025}, while HealthBench also covers assistant-facing skills such as uncertainty handling and communication.

\paragraph{Legal domain.} CAPS-Legal shows the clearest advantage in a standard-rich domain: using only public statutes and no human preference labels, it improves both 8B and 14B models over base and substantially outperforms the 10k Infinity-Instruct legal subset. The 35k HelpSteer3 baseline also does not reliably transfer to PLawBench/MSLR, especially degrading MSLR, suggesting that explicit statutory criteria provide more targeted supervision than either general preference data or legal-domain instruction data alone. CAPS-14B also slightly exceeds Qwen3-32B on MSLR (65.77 vs.\ 65.08).

\section{Analysis}
\subsection{Capability Gains Beyond Data Scale and Distillation}
\label{sec:system_eval}

\begin{figure}[ht]
\centering
\includegraphics[width=\columnwidth]{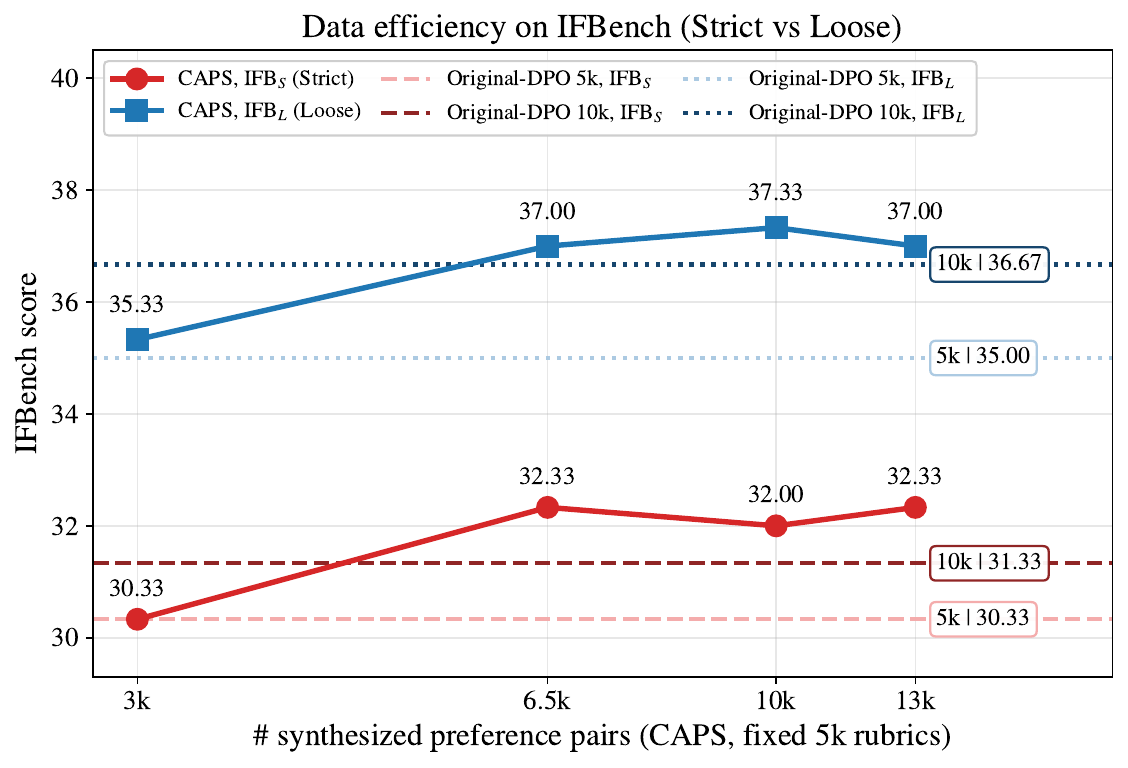}
\caption{Data-efficiency comparison on IFBench under a fixed 5k pair-level rubric budget. CAPS (red) synthesizes varying numbers of preference pairs from the same rubrics; Original-DPO baselines (gray) use 5k and 10k sampled HelpSteer3 pairs. CAPS at 6.5k pairs already exceeds the 10k Original-DPO baseline on both metrics and saturates beyond.}
\label{fig:data_efficiency}
\end{figure}

\paragraph{Data efficiency.} Under a fixed 5k HelpSteer3-rubric budget, CAPS matches the 5k Original-DPO baseline with 3k synthesized pairs and modestly outperforms the 10k Original-DPO baseline at 6.5k pairs (Figure~\ref{fig:data_efficiency}). Further scaling to 10k--13k pairs brings only marginal gains, suggesting that criterion-bound synthesis can improve data efficiency, though the gains are moderate in this setting.

\paragraph{Distillation control.} Table~\ref{tab:distillation} shows that conventional rollout-SFT distillation does not explain CAPS gains: training Qwen3-8B on responses from stronger models slightly reduces IFBench performance relative to the base model. Since Qwen3-8B is already a strong base model, direct SFT on generated responses may disrupt its instruction-following behavior; in contrast, full CAPS DPO reaches 35.00/38.33, indicating that the key signal is contrastive supervision over criterion-activated pairs.

\begin{table}[ht]
\centering
\small
\begin{tabular}{lcc}
\toprule
\textbf{Method} & \textbf{IFB$_S$} & \textbf{IFB$_L$} \\
\midrule
Qwen3-8B (Base) & 25.67 & 30.00 \\
Qwen3-235B Rollout SFT & 24.33 & 28.67 \\
Gemma-4-31B Rollout SFT & 23.33 & 26.00 \\
\textbf{CAPS DPO (Full)} & \textbf{35.00} & \textbf{38.33} \\
\bottomrule
\end{tabular}
\caption{Distillation-control baselines on IFBench. Rollout SFT denotes SFT on answers generated by Qwen3-235B or Gemma-4-31B from the original 35k HelpSteer3 queries.}
\label{tab:distillation}
\end{table}

\subsection{What Drives CAPS Gains?}

We design two complementary ablations to validate CAPS's core design. Table~\ref{tab:ablation_logic} provides a pipeline-level comparison against single-judge controls; Table~\ref{tab:ablation_component} isolates the contribution of each filtering component.

\begin{table}[t]
\centering
\small
\begin{tabular}{lcc}
\toprule
\textbf{Setting} & \textbf{IFB$_S$} & \textbf{PLaw} \\
\midrule
Qwen3-8B Base & 25.67 & 41.29 \\
\midrule
Random & 29.67 & 33.39 \\
General judge & 31.67 & 42.49 \\
Rubric judge & 31.00 & 44.10 \\
CAPS (Rubric+General Judge) & \textbf{35.00} & \textbf{47.89} \\
\bottomrule
\end{tabular}
\caption{Pipeline-level comparison. ``Random'' samples preference pairs from the rollout pool without judge-based filtering. ``General judge'' and ``Rubric judge'' use only Signal~B or Signal~A for single-signal pair selection, respectively, and therefore cannot apply the full EAT+AR verification. ``CAPS (Rubric+General Judge)'' applies the full dual-signal EAT+AR pipeline.}
\label{tab:ablation_logic}
\end{table}

\paragraph{Structured synthesis matters beyond single-judge signals.} As shown in Table~\ref{tab:ablation_logic}, random pairing improves IFB$_S$ but reduces PLawBench from 41.29 to 33.39. General criteria often encode observable requirements such as formatting or coverage, so a random direction can occasionally agree with genuine instruction-following differences. Legal responses can instead diverge in the applicable statute, exception, or doctrinal path while remaining fluent; random pairing therefore more often inverts the legally correct preference and DPO reinforces the reversed signal. Both single-judge controls improve over random selection, but each is limited by its signal: the general judge is stronger on IFBench, whereas the rubric judge is stronger on PLawBench. Full CAPS performs best in both settings (35.00 IFB$_S$, 47.89 PLaw), showing that rollout diversity alone is insufficient and that the two signals are complementary.

\begin{table}[t]
\centering
\small
\begin{tabular}{lcc}
\toprule
\textbf{Setting} & \textbf{IFB$_S$} & \textbf{PLaw} \\
\midrule
Full CAPS & 35.00 & 47.89 \\
\quad $-$ AR ranking & 33.00 & 46.95 \\
\quad $-$ AR (entire) & 31.33 & 46.65 \\
\quad $-$ EAT & 28.67 & 46.21 \\
\bottomrule
\end{tabular}
\caption{Component ablation (subtractive). Starting from Full CAPS, we progressively remove: AR-based ranking (pairs selected randomly among AR-passing candidates), AR filtering entirely (all EAT-passed pairs retained), and EAT (all criterion-context pairs advance to pair selection).}
\label{tab:ablation_component}
\end{table}

\paragraph{Each component contributes incrementally.} Table~\ref{tab:ablation_component} shows progressive degradation as AR ranking, AR filtering, and EAT are removed, consistent with their roles in Section~\ref{sec:method:eat}--\ref{sec:method:pair}. The effect is stronger on IFBench: removing EAT drops IFB$_S$ from 35.00 to 28.67, suggesting that general rubrics rely more on group-level activation checks. On PLawBench, the drops are smaller but consistent (47.89 to 46.21), likely because statutory criteria are already closely aligned with the target domain. Overall, EAT controls criterion-context learnability, while AR improves pair-level attribution and ranking.


\begin{table*}[t]
\centering
\footnotesize
\setlength{\tabcolsep}{4pt}
\renewcommand{\arraystretch}{1.05}

\begin{tabularx}{\textwidth}{
  >{\raggedright\arraybackslash}p{0.18\textwidth}
  >{\raggedright\arraybackslash}p{0.22\textwidth}
  >{\raggedright\arraybackslash}p{0.24\textwidth}
  >{\raggedright\arraybackslash}X
}
\toprule
\textbf{Check} &
\textbf{Reference setting} &
\textbf{Tested variation} &
\textbf{Downstream result} \\
\midrule

EAT threshold
& no EAT: 44.81
& $[0.2,3.0]$ to $[0.8,1.5]$
& 45.22--45.54 HealthBench \\

AR threshold
& no AR: 44.62
& $\tau_{\mathrm{AR}}=0.4$--$0.8$
& 45.30--45.50 HealthBench \\

Signal-A thresholds
& $(7,7)$: 45.30
& $(6,6)$, $(8,8)$, $(8,6)$
& 44.70--45.37 HealthBench \\

Blind-judge backbone
& DeepSeek-V3.2
& three heterogeneous judges
& $\rho=0.83$--$0.84$; top-30\% Jaccard $=1.00$ \\

Score aggregation
& median: 47.89
& no trimming
& 47.55 PLawBench \\

Criterion structure
& 6,158 single-provision pairs
& 16,001 multi-provision pairs
& 47.89 vs.\ 48.25 PLawBench \\

\bottomrule
\end{tabularx}

\caption{Summary of robustness checks. Threshold sweeps use the medical setting; aggregation and criterion-structure checks use the legal setting. The heterogeneous-judge check uses 50 stratified pairs. The multi-provision comparison changes retained data size and is therefore a scope check rather than a controlled improvement claim.}
\label{tab:robustness_summary}
\end{table*}

\paragraph{Robustness and broader criteria.} Table~\ref{tab:robustness_summary} shows that every tested EAT interval and AR threshold outperforms removing the corresponding filter, and performance varies modestly across retained settings. Replacing the blind judge largely preserves high-AR selection, while aggregation and multi-provision checks remain stable. These results support robustness but neither establish judge independence nor isolate the effect of criterion composition. Appendix~\ref{app:robustness} provides complete settings and results.

\section{Related Work}
\label{sec:related_work}
\paragraph{Preference data quality and synthesis.}
Scaling approaches~\citep{cui2024ultrafeedback,chen2024selfplay,yuan2024selfrewarding} produce large preference datasets without controlling \textit{what} differentiates chosen from rejected. CLAIR~\citep{clair2025} constrains via minimal edits, but only 43\% Jaccard similarity between intended and actual attributions. Quality-filtering work~\citep{moreless2025,air2025,sweetspot2025,pvar2025,deita2024} demonstrates curated subsets outperform full-scale data, but all apply \textit{post-hoc filtering}. CAPS \textit{constructs} attribution-traceable pairs from explicit criteria by design.

\paragraph{Rubric-guided methods.}
Rubric-based approaches span four layers: \emph{generation}~\citep{rrd2025,autorule2025,cdrrm2025}, \emph{data synthesis}~\citep{omnirrm2025,codeclm2024,rdpo2025}, \emph{reward modeling}~\citep{rar2025,openrs2025,rubicon2025}, and \emph{evaluation}~\citep{prometheus2023,sibylsense2025}. Across all four, criteria serve as evaluation standards or reward signals---never as the \textit{source specification} for constructing preference pairs. Specifically, rDPO~\citep{rdpo2025} uses rubrics as \emph{post-hoc soft labels} over already-collected pairs, and CLAIR~\citep{clair2025} controls attribution \emph{implicitly} via minimal-edit heuristics. CAPS uses criteria \emph{prospectively}---as both the generative seed for context synthesis and the explicit verification target for pair selection. No prior method separates \emph{criterion learnability} from \emph{pair attribution purity}: PVar~\citep{pvar2025} verifies prompt-level diversity but not criterion-context activation; difficulty-based selection~\citep{difficulty2025} verifies signal strength but not its source. EAT and AR fill this dual gap.

\paragraph{DPO overoptimization.}
Rank deficiency admits degenerate policies~\citep{rafailov2024dpo}, manifesting as length bias~\citep{park2024length} and reward hacking~\citep{scalinglaws2024}. Existing solutions modify the algorithm~\citep{azar2024general,ethayarajh2024kto} but leave data unchanged. CAPS instead targets the data side by localizing preference gaps to explicit criteria and reducing confounding; we do not claim or measure an increase in representation rank.

\section{Conclusion}
We introduced CAPS, a criterion-first framework that uses explicit criteria to guide preference synthesis and verifies criterion--quality alignment through EAT and AR.

Across general, medical, and legal domains, CAPS provides an effective alternative to scaling preference data alone: 20k general CAPS pairs outperform 35k HelpSteer3 pairs, and statute-derived legal supervision improves Qwen3-8B/14B by +4.9 to +6.6 points on PLawBench/MSLR without human preference labels. Future work includes dynamic criterion evolution and conflict-aware multi-criterion synthesis.

\section*{Limitations}
\paragraph{Cost and evaluation.} Dual-signal verification dominates synthesis cost: the medical pipeline uses 5.38M judge calls, with an estimated API cost of \$1.6k--\$1.9k (Appendix~\ref{app:cost}). All downstream results are from a single training run, so marginal medical gains require multi-seed validation.

\paragraph{Criteria and judges.} CAPS depends on well-specified criteria and capable judges; EAT cannot repair poor criteria, and difficult domain reasoning can make AR noisy. Both signals use DeepSeek-V3.2 with separate prompts, so their biases may remain correlated. A 50-pair heterogeneous-judge check preserves similar rankings (Appendix~\ref{app:robustness}) but does not establish independence.

\paragraph{Scope.} CAPS supports jointly applicable provisions, but the multi-provision experiment retains more data and therefore does not isolate the effect of criterion composition (Appendix~\ref{app:robustness}). Conflicts among independent objectives remain unresolved. Legal evaluation is limited to Chinese law, leaving cross-lingual and cross-jurisdictional transfer unvalidated.

\bibliography{custom}

\appendix
\section{Ethics Statement}
CAPS-Legal synthesizes preference data from publicly available legal statutes. The resulting dataset is intended solely for LLM alignment research and must not be used as a substitute for professional legal advice. No private or personally identifiable information is used in its construction.

\section{Prompt Templates}
\label{app:prompts}

Below we provide representative prompt templates for each stage of the CAPS pipeline. Placeholders in braces correspond to instantiated inputs or hyperparameters used during query generation, scoring, and filtering.
\subsection{Query Generation}

We use domain-specific generation prompts for general, medical, and legal data.

\begin{promptbox}{General Query Generation}
You are a prompt engineer. Based on the following rule, generate {num_queries} realistic user queries and the implicit constraints they require.

Rule:
{rewritten_rule}

Guidelines:
1. Queries should sound like real user requests, not meta-instructions.
2. Each query should trigger the need to follow the rule; ignoring the rule should lead to a weak answer.
3. The constraint should summarize the requirement of the rule, but must not appear verbatim in the query.
4. Vary the intent and surface form across queries.

Output format:
[
{
  "query": "realistic user question 1",
  "constraint": "corresponding implicit constraint 1"
},
{
  "query": "realistic user question 2",
  "constraint": "corresponding implicit constraint 2"
}
]
\end{promptbox}

\begin{promptbox}{Medical Query Generation}
You are generating a self-contained medical QA training query from a rubric.

Input rubric:
{rubric}

Requirements:
1. Treat the rubric as the only task input. Do not rely on any hidden original question or reference answer.
2. Produce exactly one medically coherent English training sample that would naturally elicit an answer satisfying the rubric.
3. Infer a realistic clinical scenario, exam stem, or decision-making setup from the rubric itself.
4. Add concise but sufficient clinical context when needed so the sample is self-contained.
5. Do not mention "rubric", "criteria", "reference answer", or any annotation language in the final sample.
6. Question should be the final user-facing medical question.
7. Query should be the exact text that will be sent to rollout models. If context is non-empty, query should combine context and question naturally.
8. answer_focus should briefly summarize the key medical points the ideal answer must cover.
9. Output JSON only:
{{
  "context": "...",
  "question": "...",
  "query": "...",
  "answer_focus": "short note on what the answer must get right"
}}
\end{promptbox}

\begin{promptbox}{Law Query Generation}
You are a Chinese legal-case data synthesis expert. Based on the given legal provision, generate {variants_per_article} groups of "case context + question" samples in Chinese, for testing whether a model can provide correct legal analysis in a
complete factual scenario.

Requirements:
1. Output JSON only. Do not output Markdown or explanations.
2. Each variant must contain one constraint, one context, and one question.
3. The sample should resemble realistic legal training data: the context should be a complete case narrative, and the question should be a clear legal question. Do not write it as a short chat-style consultation.
4. The context must contain sufficient factual density, preferably 250-500 Chinese characters, including multiple elements such as parties, timeline, key conduct, contracts/evidence/communications, administrative or judicial progress,
counterarguments, dispute focus, losses, or risks.
5. The question should be short and clear. It may be judgment/choice-style or explanation/analysis-style; avoid fixed templates.
6. Across every {variants_per_article} variants, include both judgment/choice-style questions and explanation/analysis-style questions.
7. The context and question must not explicitly repeat or closely paraphrase the abstract wording of the rubric/provision. Do not include meta-expressions such as "explain this statute", "according to this article", or "whether it satisfies this
element".
8. Use concrete facts to naturally elicit the target legal norm, rather than inserting normative keywords into the question.
9. Each constraint must be a key factual condition or boundary condition explicitly present in the context. The constraint is an internal audit field and should not look copied into the context/question.
10. The core dispute should be primarily determined by this provision, or this provision should be an indispensable premise. If the provision is merely background, the sample is invalid.
11. The {variants_per_article} variants should differ substantially, covering different parties, factual boundaries, liability forms, exceptions, evidence, deadlines, good faith/bad faith, recourse, or exemption angles.
12. If the provision comes from criminal law, civil procedure, labor law, personal information protection, or other domains, the scenario must naturally fall into that domain.
13. You may add a label field, such as "personal life", "labor dispute", "business operation", "platform transaction", or "litigation procedure".

Output JSON schema:
{
"variants": [
  {
    "variant_id": 1,
    "label": "short Chinese scenario label",
    "context": "complete Chinese case context, 250-500 characters",
    "constraint": "single key constraint triggered by this query",
    "question": "final Chinese question for the model",
    "answer_focus": "key legal points the ideal answer must cover"
  }
]
}

Article number:
{article_no}

Legal provision:
{rubric}
\end{promptbox}

\subsection{Rubric-Bound Score (Signal A)}

\begin{promptbox}{Rubric-Bound Score (Signal A): General Domain}
You are a strict rubric evaluator.

Score how well the model response satisfies the given rubric and explicit constraint for the user query.

User Query:
{query}

Rubric:
{rubric}

Constraint:
{constraint}

Model Response:
{response}

Return JSON only:
{{
  "score": an integer from 1 to 10,
  "reason": "short reason"
}}

Score guide:
1-2  = clearly violates the rubric or constraint.
3-4  = weakly satisfies the rubric with major omissions.
5-6  = partially satisfies the rubric with important gaps.
7-8  = mostly satisfies the rubric and constraint with minor issues.
9-10 = fully satisfies the rubric and constraint.
\end{promptbox}

\begin{promptbox}{Rubric-Bound Score (Signal A): Medical Domain}
You are a strict medical-answer evaluator.

Score how well the model response satisfies the given medical rubric for the user query.

User Query:
{query}

Rubric:
{rubric}

Model Response:
{response}

Return JSON only:
{{
  "score": an integer from 1 to 10,
  "reason": "short reason"
}}

Score guide:
1-2  = clearly violates or misses the rubric.
3-4  = weakly satisfies the rubric with major omissions.
5-6  = partially satisfies the rubric with important gaps.
7-8  = mostly satisfies the rubric with minor issues.
9-10 = fully satisfies the rubric.
\end{promptbox}

\begin{promptbox}{Rubric-Bound Score (Signal A): Legal Domain}
You are a strict legal-answer evaluator.

Score how well the model response satisfies the given legal rubric for the user query.

User Query:
{query}

Rubric:
{rubric}

Model Response:
{response}

Return JSON only:
{{
  "score": an integer from 1 to 10,
  "reason": "short reason"
}}

Score guide:
1-2  = clearly violates or misses the rubric.
3-4  = weakly satisfies the rubric with major omissions.
5-6  = partially satisfies the rubric with important gaps.
7-8  = mostly satisfies the rubric with minor issues.
9-10 = fully satisfies the rubric.
\end{promptbox}

\subsection{General Blind Score (Signal B)}

\begin{promptbox}{General Blind Score (Signal B): General Domain}
You are a strict answer-quality evaluator.

Score the overall quality of the model response to the user query. Evaluate whether the response is relevant, correct, helpful, complete, and coherent.

User Query:
{query}

Model Response:
{response}

Return JSON only:
{{
  "score": an integer from 1 to 10,
  "reason": "short reason"
}}

Score guide:
1-2  = does not answer or is clearly wrong.
3-4  = mostly unhelpful, vague, or seriously flawed.
5-6  = partially useful but incomplete or weak.
7-8  = useful and coherent with minor issues.
9-10 = directly answers the request with strong quality.
\end{promptbox}

\begin{promptbox}{General Blind Score (Signal B): Medical Domain}
You are a strict medical-answer evaluator.

Score the overall quality of the model response to the medical user query. Evaluate whether the response is medically relevant, correct, helpful, complete, safe, and well-reasoned.

User Query:
{query}

Model Response:
{response}

Return JSON only:
{{
  "score": an integer from 1 to 10,
  "reason": "short reason"
}}

Score guide:
1-2  = does not answer, is medically wrong, or unsafe.
3-4  = mostly unhelpful, vague, or seriously flawed.
5-6  = partially useful but incomplete or weak.
7-8  = useful and medically coherent with minor issues.
9-10 = directly answers the medical question with strong reasoning.
\end{promptbox}

\begin{promptbox}{General Blind Score (Signal B): Legal Domain}
You are a strict legal-answer evaluator.

Score the overall quality of the model response to the legal user query. Evaluate whether the response is legally relevant, correct, helpful, complete, and well-reasoned.

User Query:
{query}

Model Response:
{response}

Return JSON only:
{{
  "score": an integer from 1 to 10,
  "reason": "short reason"
}}

Score guide:
1-2  = does not answer or is legally wrong.
3-4  = mostly unhelpful, vague, or seriously flawed.
5-6  = partially useful but incomplete or weak.
7-8  = useful and legally coherent with minor issues.
9-10 = directly answers the legal issue with strong reasoning.
\end{promptbox}
\section{CAPS-Legal Profile}
\label{app:caps_legal_profile}

\begin{figure*}[t]
\centering
\includegraphics[width=\textwidth]{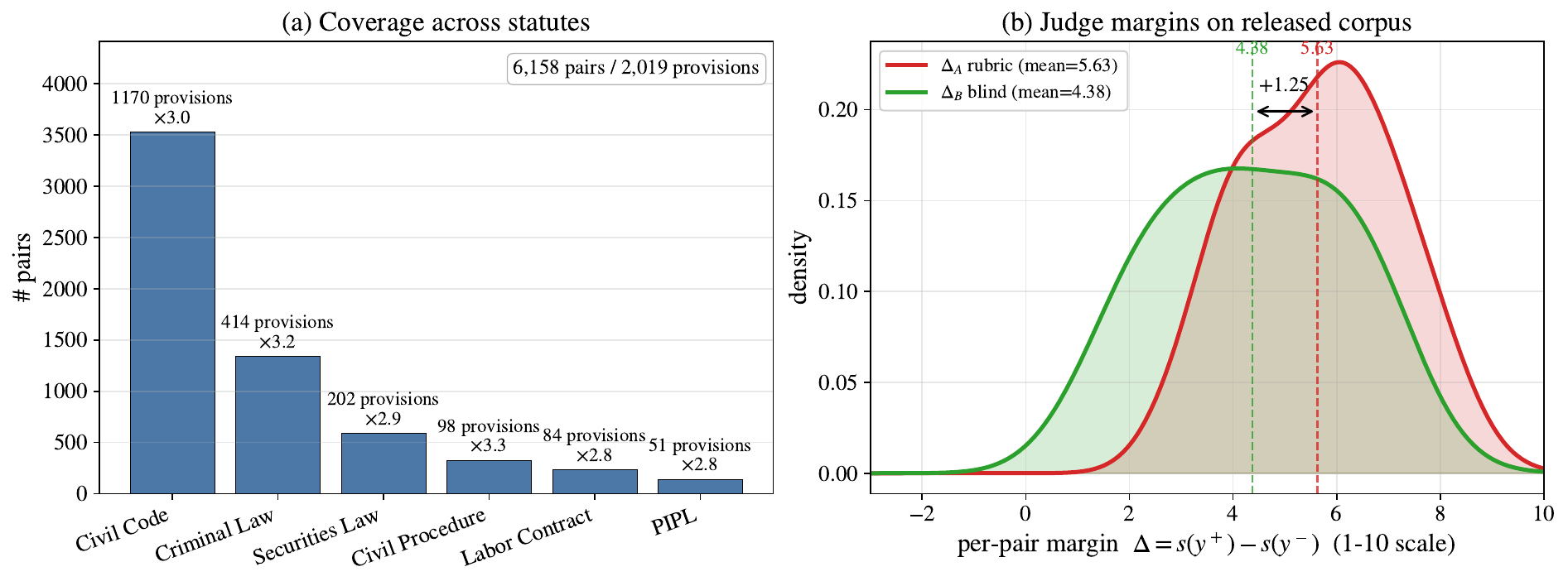}
\caption{CAPS-Legal corpus diagnostics. \textbf{(a)}~Coverage across six statutes. \textbf{(b)}~Per-pair margin distributions under the rubric judge ($\Delta_A$) and the blind judge ($\Delta_B$); the $+1.25$ shift reflects the rubric judge's stronger sensitivity to the curated pairs.}
\label{fig:legal_data_diagnostics}
\end{figure*}

CAPS-Legal contains 6{,}158 preference pairs derived from 2{,}019 distinct provisions ($\sim$3 pairs/rule) across six Chinese statutes---Civil Code, Criminal Law, Securities Law, Civil Procedure Law, Labor Contract Law, and Personal Information Protection Law (Fig.~\ref{fig:legal_data_diagnostics}a). On this filtered corpus, rubric-conditioned margins $\Delta_A$ are right-shifted from blind margins $\Delta_B$ by $+1.25$ on the 1--10 scale (Fig.~\ref{fig:legal_data_diagnostics}b), indicating that the rubric judge perceives a substantially larger preference gap than the blind judge on the same retained pairs.

\section{Hyperparameters}
\label{app:hyperparams}

Table~\ref{tab:hyperparams} lists all hyperparameters used in the CAPS pipeline and DPO training.

\begin{table}[ht]
\centering
\footnotesize
\begin{tabular}{lcp{3cm}}
\toprule
\textbf{Parameter} & \textbf{Value} & \textbf{Description} \\
\midrule
\multicolumn{3}{l}{\textit{EAT Filtering}} \\
$\tau_\text{min}$ & 0.5 & EAT discriminability lower bound (pass rate floor) \\
$\tau_\text{max}$ & 2.0 & EAT discriminability upper bound (pass rate ceiling) \\
$K_\text{ctx}$ & 10 / 2 & Contexts per criterion for General/Legal / Medical \\
$K_\text{roll}$ & 20 & Responses per context (one per model-temperature combination) \\
\midrule
\multicolumn{3}{l}{\textit{Dual-Signal Verification}} \\
$K_\text{vote}$ & 3 & Samples per response per signal; aggregation is the median \\
$\tau^+$ & 7.0 & Signal A score upper threshold (satisfy group, on 1--10 scale) \\
$\tau^-$ & 7.0 & Signal A score lower threshold (violate group, on 1--10 scale) \\
$\tau_\text{AR}$ & 0.6 & Attribution Ratio ranking threshold for pair retention \\
\midrule
\multicolumn{3}{l}{\textit{Pair Construction}} \\
$q_\text{min}$ & 6.0 & Minimum quality score for chosen responses \\
$\Delta_\text{min}$ & 1.0 & Minimum quality margin (chosen $-$ rejected) \\
$\Delta_\text{max}$ & 5.0 & Maximum quality margin (avoids trivial pairs) \\
$K_\text{rep}$ & 2 & Maximum pairings per individual response \\
$K_\text{pair}$ & 5 & Maximum preference pairs per (context, criterion) \\
\midrule
\multicolumn{3}{l}{\textit{DPO Training}} \\
$\beta$ & 0.1 & DPO temperature parameter \\
Learning rate & $2 \times 10^{-6}$ & With cosine decay schedule \\
Batch size & 32 & Global batch size \\
Epochs & 3 & Total training epochs \\
Max length & 4096 & Maximum sequence length \\
Warmup ratio & 0.05 & Linear warmup proportion \\
\bottomrule
\end{tabular}
\caption{Complete hyperparameter specification for the CAPS pipeline and DPO training.}
\label{tab:hyperparams}
\end{table}

\section{EAT Details}
\label{app:eat}

\paragraph{Cohen's $d$ interpretation.} Table~\ref{tab:cohens_d} provides standard interpretation benchmarks~\citep{cohen1977statistical} and their mapping to CAPS decisions.

\begin{table}[ht]
\centering
\small
\begin{tabular}{@{}lcl@{}}
\toprule
\textbf{Effect Size} & \textbf{Cohen's $d$} & \textbf{CAPS Decision} \\
\midrule
Negligible & $d < 0.2$ & No activation; discarded \\
Small & $0.2 \leq d < 0.5$ & Weak activation; discarded \\
Medium & $0.5 \leq d < 0.8$ & Moderate activation; retained \\
Large & $0.8 \leq d < 1.2$ & Strong activation; retained \\
Very large & $1.2 \leq d \leq 2.0$ & Very strong; retained \\
Extreme & $d > 2.0$ & Trivial distinction; discarded \\
\bottomrule
\end{tabular}
\caption{Cohen's $d$ interpretation benchmarks~\citep{cohen1977statistical} and corresponding CAPS decisions. The retained range $[\tau_{\min}, \tau_{\max}] = [0.5, 2.0]$ spans medium through very large effect sizes.}
\label{tab:cohens_d}
\end{table}

\paragraph{Upper bound evidence.} Four independent lines support $\tau_{\max} = 2.0$:
\begin{itemize}[nosep,leftmargin=*]
    \item \citet{air2025}: pairs with $\Delta \geq 4$ (corresponding to $d > 2.0$) degrade reward accuracy by 5.42 points.
    \item \citet{sweetspot2025}: optimal DPO at moderate margins ($\mu \pm 2\sigma$), with sharp degradation at extremes.
    \item \citet{difficulty2025}: ``too-easy'' pairs provide less per-sample signal than moderate pairs.
    \item \citet{moreless2025}: trivially distinguishable pairs are preferentially exploited during reward hacking.
\end{itemize}

\paragraph{Detailed case analysis of extreme $d$ values.} Since Cohen's $d = (\bar{s}^{(B),+} - \bar{s}^{(B),-})/s_p$, extreme values arise from either the numerator (group mean difference) or denominator (pooled standard deviation):

\smallskip\noindent\textit{$d < \tau_{\min}$ --- Scenario 1 (small numerator):} $\bar{s}^{(B),+} \approx \bar{s}^{(B),-}$. The criterion has no discriminative power in this context---satisfying or violating $c$ produces indistinguishable quality. Example: ``cite legal sources'' applied to a creative writing prompt. \textbf{Action:} discard.

\smallskip\noindent\textit{$d < \tau_{\min}$ --- Scenario 2 (large denominator):} The mean difference is non-trivial, but $s_p$ is inflated by within-group variance. Two independent sources: (a)~rollout heterogeneity---responses from different model scales (8B/14B/235B) create high intra-group variance; (b)~judge scoring instability---the same response receives different scores across calls. The real signal is masked. \textbf{Action:} discard at group level, though AR may recover individual high-attribution pairs.

\smallskip\noindent\textit{$d > \tau_{\max}$ --- Scenario 3 (large numerator):} $\bar{s}^{(B),+} \gg \bar{s}^{(B),-}$ with normal $s_p$. The groups barely overlap, indicating a trivially obvious distinction. Example: ``respond in target language'' applied to an English prompt---violations produce off-language gibberish. Training on such pairs encourages reward hacking on surface cues. \textbf{Action:} discard.

\smallskip\noindent\textit{$d > \tau_{\max}$ --- Scenario 4 (near-zero denominator):} $s_p \to 0$ due to few samples with coarse-grained judge scoring. Example: $\mathbf{s}^{(B),+} = \{7,7,7\}$, $\mathbf{s}^{(B),-} = \{6,6,6\}$ yields $d \to \infty$ despite only 1-point absolute difference. This is a statistical artifact from degenerate variance estimation, not genuine activation. \textbf{Action:} discard.

\paragraph{Complementary verification axes.} Table~\ref{tab:verification_axes} contrasts EAT with related verification mechanisms.

\begin{table}[ht]
\centering
\footnotesize
\begin{tabular}{@{}lll@{}}
\toprule
\textbf{Method} & \textbf{Verifies} & \textbf{Property} \\
\midrule
PVar & Prompt & Diverse responses \\
Difficulty & Pair & Appropriate gap \\
EAT (ours) & Crit.$\times$Ctx & Discriminative activation \\
\bottomrule
\end{tabular}
\caption{Three complementary verification axes addressing distinct quality dimensions.}
\label{tab:verification_axes}
\end{table}

\section{Pair Quality Constraints}
\label{app:pair_constraints}

Beyond the AR threshold, we impose four constraints grounded in empirical findings:

\begin{table}[ht]
\centering
\footnotesize
\begin{tabular}{@{}p{1.8cm}p{1.5cm}p{3.2cm}@{}}
\toprule
\textbf{Constraint} & \textbf{Value} & \textbf{Support} \\
\midrule
Directional & $\Delta_A\!>\!0, \Delta_B\!>\!0$ & Eliminates paradoxical pairs \\
Chosen quality & $s^{(B)}_{y^+} \!\geq\! 6$ & +9.35\% accuracy~\citep{air2025} \\
Margin & $\Delta_B \!\in\! [1,5]$ & Optimal at $\mu\!\pm\!2\sigma$~\citep{sweetspot2025} \\
Diversity & $K_{\text{rep}}\!=\!2$ & $-$1.67 without~\citep{rdpo2025} \\
\bottomrule
\end{tabular}
\caption{Pair quality constraints with empirical justification.}
\label{tab:pair_constraints}
\end{table}

\begin{itemize}[nosep,leftmargin=*]
    \item \textbf{Directional consistency:} $\Delta_A > 0$ and $\Delta_B > 0$, ensuring chosen dominates rejected on both criterion compliance and holistic quality.
    \item \textbf{Chosen quality:} $s^{(B)}_{y^+} \geq 6.0$, ensuring adequate overall quality. High-quality chosen responses improve reward model accuracy~\citep{air2025}.
    \item \textbf{Margin bounds:} $\Delta_B \in [1.0, 5.0]$, avoiding noisy pairs (too small) and trivial pairs (too large).
    \item \textbf{Diversity cap:} Each response appears in at most $K_{\text{rep}} = 2$ selected pairs, preventing a small set of responses from dominating the data.
\end{itemize}

\section{Multi-Model Risk Mitigation}
\label{app:multimodel}

\citet{moreless2025} demonstrate that naively mixing multi-model data induces reward hacking (ASR from 0.12 to 0.51). CAPS mitigates this through three layers:
\begin{enumerate}[nosep,leftmargin=*]
    \item \textbf{Criterion-bound grouping:} Pairs are formed based on criterion satisfaction status, not model identity.
    \item \textbf{AR filtering:} Stylistic differences inflate $\Delta_B$ without proportionally increasing $\Delta_A$, yielding low AR scores for confound-dominated pairs.
    \item \textbf{EAT group-level gate:} Ensures surface-level model differences alone cannot produce passing criterion-context pairs.
\end{enumerate}

\section{Attribution Ratio Computation}
\label{app:ar_computation}

Given a candidate pair $(y^+, y^-)$ for context $x$ and criterion $c$:

\begin{align}
\Delta_A &= s^{(A)}_{y^+} - s^{(A)}_{y^-} \\
\Delta_B &= s^{(B)}_{y^+} - s^{(B)}_{y^-} \\
\text{AR} &= \frac{\Delta_A}{\max(\Delta_A, \Delta_B) + \varepsilon}
\end{align}

where $s^{(A)}_k = J^{(A)}(x, y_k, c)$ is the rubric-bound score (Signal~A) and $s^{(B)}_k = J^{(B)}(x, y_k)$ is the general blind score (Signal~B). The AR is bounded in $[0, 1]$ by construction: when $\Delta_A \geq \Delta_B$, AR $= \Delta_A/(\Delta_A + \varepsilon) \approx 1$ for $\Delta_A \gg \varepsilon$; when $\Delta_A < \Delta_B$, AR $= \Delta_A / (\Delta_B + \varepsilon) < 1$ (non-criterion factors dominate). Pairs are retained when $\text{AR} \geq \tau_\text{AR} = 0.6$, which serves as a ranking threshold to select pairs where the criterion mediates a substantial proportion of the quality gap. We additionally require $\Delta_A > 0$ and $\Delta_B > 0$ to ensure both that criterion compliance differs and that general quality supports the preference direction.

\subsection{Controlled Data-Efficiency Check}
\label{app:data_efficiency}

To complement the main setup in Section~\ref{sec:experiments}, Table~\ref{tab:data_efficiency} isolates the effect of synthesized pair count under a fixed 5k-rubric budget. With comparable or fewer training pairs, CAPS matches or exceeds Original-DPO on IFBench, suggesting that rubric-conditioned synthesis can use criterion sources efficiently.

\begin{table}[ht]
\centering
\small
\begin{tabular}{lcccc}
\toprule
\textbf{Method} & \textbf{\# Rubrics} & \textbf{\# Pairs} & \textbf{IFB$_S$} & \textbf{IFB$_L$} \\
\midrule
Original-DPO & 5.0k & 5.0k & 30.33 & 35.00 \\
Original-DPO & 10.0k & 10.0k & 31.33 & 36.67 \\
\midrule
CAPS (Ours) & 5.0k & 3.0k & 30.33 & 35.33 \\
CAPS (Ours) & 5.0k & 6.5k & \textbf{32.33} & 37.00 \\
CAPS (Ours) & 5.0k & 10.0k & 32.00 & \textbf{37.33} \\
CAPS (Ours) & 5.0k & 13.0k & \textbf{32.33} & 37.00 \\
\bottomrule
\end{tabular}
\caption{Data-efficiency comparison on IFBench under controlled rubric budgets. Original-DPO uses sampled HelpSteer3 preference pairs, while CAPS synthesizes preference pairs from a fixed 5k subset of pair-level rubrics extracted from HelpSteer3.}
\label{tab:data_efficiency}
\end{table}

\section{Robustness and Structured Criteria}
\label{app:robustness}

\paragraph{Threshold sensitivity.} We vary one component at a time in the medical setting. The EAT sweep fixes $\tau=(7,7)$ and $\tau_{\mathrm{AR}}=0.6$; the AR sweep fixes EAT to $[0.5,2.0]$ and $\tau=(7,7)$; and the Signal-A sweep fixes EAT to $[0.5,2.0]$ and $\tau_{\mathrm{AR}}=0.6$.

\begin{table}[ht]
\centering
\small
\begin{tabular}{llc}
\toprule
\textbf{Sweep} & \textbf{Setting} & \textbf{HealthBench} \\
\midrule
Baseline & Qwen3-8B & 41.53 \\
Baseline & HelpSteer3 & 45.19 \\
Baseline & RaR-Med. & 41.79 \\
\midrule
EAT & none & 44.81 \\
EAT & $[0.2,3.0]$ & 45.22 \\
EAT & $[0.5,2.0]$ & 45.30 \\
EAT & $[0.8,1.5]$ & 45.54 \\
$\tau_{\mathrm{AR}}$ & none & 44.62 \\
$\tau_{\mathrm{AR}}$ & 0.4 & 45.37 \\
$\tau_{\mathrm{AR}}$ & 0.6 & 45.30 \\
$\tau_{\mathrm{AR}}$ & 0.8 & 45.50 \\
Signal-A $\tau$ & $(6,6)$ & 44.96 \\
Signal-A $\tau$ & $(7,7)$ & 45.30 \\
Signal-A $\tau$ & $(8,8)$ & 44.70 \\
Signal-A $\tau$ & $(8,6)$ & 45.37 \\
\bottomrule
\end{tabular}
\caption{Medical-domain sensitivity analysis. Every tested EAT interval and AR threshold outperforms removing the corresponding filter; variation across retained settings is modest.}
\label{tab:sensitivity}
\end{table}

\paragraph{Heterogeneous blind judges.} We keep Signal~A on DeepSeek-V3.2 and re-score Signal~B with three heterogeneous closed-source judges on 50 stratified pairs spanning General, Medical, and Legal. Table~\ref{tab:judge_robustness} compares the resulting AR values and selections with the original pipeline. The small sample does not prove independence, but high-AR selection is largely preserved.

\begin{table}[ht]
\centering
\footnotesize
\setlength{\tabcolsep}{3pt}
\begin{tabular}{lccc}
\toprule
\textbf{Metric} & \textbf{GPT-5.4-mini} & \textbf{Gemini-3.5} & \textbf{Claude-4.6} \\
\midrule
Spearman $\rho$ & 0.83 & 0.84 & 0.84 \\
AR agreement & 0.78 & 0.64 & 0.79 \\
Top-30\% Jaccard & 1.00 & 1.00 & 1.00 \\
Reversal rate & 8\% & 2\% & 6\% \\
\bottomrule
\end{tabular}
\caption{Robustness to replacing the blind judge. Model names abbreviate gpt-5.4-mini, gemini-3.5-flash, and claude-sonnet-4-6.}
\label{tab:judge_robustness}
\end{table}

\paragraph{Aggregation robustness.} With $K_{\mathrm{vote}}=3$, trimming the highest and lowest scores is equivalent to taking the median. Removing trimming changes PLawBench from 47.89 to 47.55, indicating that aggregation stabilizes scoring but is not the main source of the gain.

\paragraph{Multi-provision criteria.} We use an LLM to group provisions that may jointly govern a case or interact through conditions, exceptions, and dependencies, producing an approximately 68k-entry criterion bank. Applying the same CAPS configuration retains 16,001 pairs and obtains 48.25 PLawBench, compared with 6,158 pairs and 47.89 for the single-provision setup. Because the retained data sizes differ, this is evidence that CAPS can operate with structured criteria, not a controlled claim that composition itself improves performance. Conflicts among independent objectives remain outside the current scope.

\section{Infinity-Instruct legal baseline}
\label{app:inf_inst_legal}
To construct an external-data baseline for the legal domain, we use the BAAI Infinity-Instruct collection and rely on its native metadata tags to identify legal examples, rather than applying an additional LLM-based classifier. We first convert the original parquet shards into a JSONL corpus, and then filter examples using the dataset-provided \texttt{label} fields.

We apply a strict legal-tag filter using the dataset-provided ability labels. 
An example is retained if either its Chinese or English ability tag maps to one of the following legal-specific categories: 
\textit{legal knowledge}, \textit{understanding laws and regulations}, \textit{contract understanding}, and \textit{legal application}.
We further restrict the language field to Chinese (\texttt{langdetect = zh-cn}), yielding a strict Chinese legal subset from Infinity-Instruct.

The converted Infinity-Instruct \texttt{7M} corpus contains 7,449,106 examples. Applying the strict legal-tag and Chinese-language filters yields 10,519 (10k) examples, which we convert into the ShareGPT/SFT format. We denote this baseline as \textsc{Inf-Inst Legal}. Unlike CAPS, which constructs verified preference pairs from explicit legal criteria, this baseline uses human-authored Infinity-Instruct conversations selected by native legal metadata.

\section{Reproducibility, Artifacts, and Responsible Use}
\label{app:reproducibility}
\paragraph{Artifact licenses and intended use.}
We use publicly available source datasets, benchmarks, model artifacts, and legal materials under their respective licenses and terms. CAPS-Legal is intended for research on LLM alignment and preference learning, not for legal decision-making or as a substitute for professional advice.

For third-party artifacts used in data construction and evaluation, including HelpSteer3, RaR-Medicine, Infinity-Instruct, IFEval, IFBench, AlpacaEval, Creative Writing v3, HealthBench, PLawBench, and MSLR, we cite the original creators and follow their respective licenses or terms of use. When an artifact specifies research-only or non-commercial use, our use is restricted to research evaluation and model development within that scope.

\paragraph{Computational resources.}
All training-related experiments were conducted on a single server equipped with 8 NVIDIA H800 GPUs. This includes SFT and DPO training for the Qwen3-8B and Qwen3-14B checkpoints, as well as the corresponding ablation and data-efficiency runs. Data synthesis and LLM-judge scoring were performed through API-based model calls, while local training and evaluation jobs were run on the H800 server.

\paragraph{API cost estimate.}
\label{app:cost}
For the medical pipeline, 22,408 criteria with two queries each produce 44,816 query generations. With $K_{\mathrm{roll}}=20$, this yields 896,320 rollout responses; scoring each response with two signals and $K_{\mathrm{vote}}=3$ yields 5,377,920 judge calls. Using OpenRouter prices of \$0.2145/M input and \$0.3218/M output tokens for DeepSeek-V3.2, and \$0.117/M input and \$0.455/M output tokens for Qwen3-8B, we estimate:

\begin{table}[ht]
\centering
\footnotesize
\begin{tabular}{lrc}
\toprule
\textbf{Stage} & \textbf{Volume} & \textbf{Cost} \\
\midrule
Query synthesis & 44,816 & \$11--\$13 \\
Rollout generation & 896,320 & \$309--\$513 \\
Dual-signal verification & 5,377,920 & \$1,240--\$1,327 \\
\midrule
Total & -- & \$1.6k--\$1.9k \\
\bottomrule
\end{tabular}
\caption{Estimated medical data-construction cost before filtering.}
\label{tab:api_cost}
\end{table}

The estimate assumes approximately 1k input and 100--200 output tokens per query synthesis call, 1k input and 500--1,000 output tokens per rollout, and 1k input and 50--100 output tokens per judge call. The pipeline ultimately retains approximately 2k medical preference pairs.

\paragraph{Implementation details.}
We implement training with LLaMA-Factory and use the DPO/SFT configurations in Appendix~\ref{app:hyperparams}. Unless otherwise specified, DPO uses $\beta=0.1$, learning rate $2\times10^{-6}$, batch size 32, and 3 epochs. For response generation and rollout, we use the model-temperature combinations described in Section~3.1, with temperatures $\{0.5,0.7,0.9,1.0\}$ and $K_{\mathrm{roll}}=20$ candidates per context.

For evaluation, we use the official or publicly released implementations of IFEval, IFBench, AlpacaEval, Creative Writing v3, HealthBench, PLawBench, and MSLR whenever available. We keep benchmark-specific default settings unless otherwise stated. For LLM-judge-based scoring, both rubric-bound and blind judges use DeepSeek-V3.2 in non-thinking mode with $K_{\mathrm{vote}}=3$ stochastic samples and trimmed-mean aggregation. All random sampling steps use fixed random seeds when constructing fixed-size comparison sets.
\section{Cross-Domain Data Examples}
\label{app:data_examples}

We show one representative preference pair from each CAPS domain. In every case $\Delta_A > \Delta_B$ (AR$\,=\,1.0$), indicating that the preference is driven by criterion compliance rather than surface quality.

\begin{datacase}{General \hfill\normalfont $\Delta_A\!=\!4,\ \Delta_B\!=\!1$}

\casefield{Criterion}
{A better answer should provide multiple varied and accurate rephrased alternatives that maintain the original meaning, demonstrating completeness and creativity, rather than offering a single, potentially altered version with limited usefulness.}

\tcbline

\casefield{Synthesized Query}
{Give me a few different ways to say ``thank you for your help,'' all with the same meaning. I need more than one choice, and they must all be right. Show me a full set of options, not just one changed version. Use only simple, everyday words.}

\tcbline

\chosen{$s^{(A)}\!=\!9,\ s^{(B)}\!=\!10$}
{Sure! Here are several simple and correct ways to say ``thank you for your help,'' all with the same meaning:
\begin{itemize}[nosep,leftmargin=1.2em]
\item Thanks for helping me.
\item I really appreciate your help.
\item Thank you for giving me a hand.
\item I'm grateful for your help.
\item Thanks a lot for helping out.
\item I appreciate you helping me.
\item Thank you for being so helpful.
\item Many thanks for your support.
\item I'm so thankful you helped.
\item Big thanks for your help.
\end{itemize}
All of these are natural, polite, and commonly used in everyday conversation.}

\tcbline

\rejected{$s^{(A)}\!=\!5,\ s^{(B)}\!=\!9$}
{Here are several simple ways to say ``thank you for your help'':
\begin{itemize}[nosep,leftmargin=1.2em]
\item Thanks for your help.
\item Thanks for helping me.
\item I appreciate your help.
\item Thank you for the help.
\item Thanks a lot for helping.
\end{itemize}}

\tcbline

\casefield{Attribution}
{The blind judge rates both highly (10 vs 9)---both are polite, well-formatted lists. But the rubric judge detects that the rejected version's five options are near-duplicates (all minor rearrangements of ``thanks/help''), failing the criterion's call for \emph{variety and completeness}. The chosen version offers genuinely distinct phrasings (gratitude, appreciation, support, hand). \textbf{Signal: criterion-specific variety $>$ surface fluency.}}

\end{datacase}

\begin{datacase}{Medical \hfill\normalfont $\Delta_A\!=\!8,\ \Delta_B\!=\!2$}

\casefield{Criterion}
{Must identify omega-3 fatty acids as the medication that does not increase myopathy risk, and explain that omega-3s lower triglycerides without muscle-related side effects---unlike statins, fibrates, or niacin.}

\tcbline

\casefield{Synthesized Query}
{A 60-year-old female with hypertriglyceridemia (fasting TG 500 mg/dL) and low HDL (28 mg/dL) is evaluated for pharmacotherapy. She is physically active and concerned about muscle-related side effects. No history of myopathy. Which medication is considered safe regarding myopathy risk?}

\tcbline

\chosen{$s^{(A)}\!=\!9,\ s^{(B)}\!=\!8$}
{Omega-3 fatty acids are considered safe regarding myopathy risk for this patient. They lower triglycerides by reducing hepatic VLDL production and can modestly raise HDL cholesterol, all without increasing muscle-related side effects. This contrasts with statins, which inhibit HMG-CoA reductase and can cause myopathy, and with fibrates, which carry a dose-dependent myopathy risk especially in combination therapy. Omega-3s act through a distinct lipid-lowering pathway with no known skeletal muscle toxicity, making them an appropriate choice for a physically active patient concerned about muscle side effects.}

\tcbline

\rejected{$s^{(A)}\!=\!1,\ s^{(B)}\!=\!6$}
{Fenofibrate is considered safe regarding myopathy risk for this patient. It is associated with a lower risk compared to other fibrates like gemfibrozil, and it does not significantly increase muscle-related side effects. It is generally well-tolerated in patients without a history of myopathy, and is effective at lowering triglycerides and modestly raising HDL.}

\tcbline

\casefield{Attribution}
{The blind judge gives the rejected response moderate credit (6)---the explanation is fluent and partially accurate. But the rubric judge assigns the minimum ($s^{(A)}\!=\!1$) because fenofibrate \emph{does} carry myopathy risk, especially when combined with statins. \textbf{Signal: factual correctness on the target criterion is non-negotiable, regardless of fluency.}}

\end{datacase}

\begin{datacase}{Legal \hfill\normalfont $\Delta_A\!=\!4,\ \Delta_B\!=\!2$}

\casefield{Criterion (Civil Code Art.~104)}
{When the property of a non-legal-person organization is insufficient to settle its debts, the investor or founder bears \emph{unlimited} liability, unless otherwise provided by law.}

\tcbline

\casefield{Synthesized Query (paraphrased from Chinese)}
{Painter Chen runs ``Ink-Charm Studio,'' an unregistered personal studio. He borrows 200{,}000 RMB from a gallery in the studio's name. Before repayment, he privately transfers the studio's main assets (paintings worth 150{,}000 RMB) to a friend to offset personal debts. When the loan comes due, the studio has only 10{,}000 RMB left. The gallery sues Chen personally for the full amount. Chen argues the debt is the studio's, so liability should be capped at studio assets. Can the gallery demand Chen personally repay the full 200{,}000 RMB?}

\tcbline

\chosen{$s^{(A)}\!=\!9,\ s^{(B)}\!=\!9$}
{Yes, the gallery can demand Chen personally repay the full 200{,}000 RMB. Since ``Ink-Charm Studio'' was never registered as a legal person, it qualifies under Chinese civil law as a \emph{non-legal-person organization}. Civil Code Art.~104 directly provides that when such an organization's property is insufficient to settle its debts, the founder bears unlimited liability. The long-standing financial commingling and the below-value asset transfer further confirm that studio assets cannot be cleanly separated from Chen's personal assets. As founder, Chen must therefore repay the outstanding 190{,}000 RMB shortfall with his personal property.}

\tcbline

\rejected{$s^{(A)}\!=\!5,\ s^{(B)}\!=\!7$}
{Yes, the gallery can require Chen to personally repay the debt. Chen's personal finances and those of the studio have long been commingled, and he transferred studio assets at below-market value clearly to evade debt---this constitutes an abuse of the corporate form. Under Company Law Art.~20(3), where a shareholder abuses the independent legal personality of the company and limited liability to evade debts and seriously harms creditors, the shareholder shall be jointly liable for the company's debts. Chen's behavior clearly harms the creditor, so the gallery may demand personal repayment.}

\tcbline

\casefield{Attribution}
{Both responses reach the same correct conclusion and both read fluently---the blind judge accordingly rates them close (9 vs 7). But the rejected response invokes the \emph{wrong legal doctrine}: ``piercing the corporate veil'' applies to registered companies, not to unregistered non-legal-person organizations. The applicable rule is Civil Code Art.~104, which directly imposes unlimited liability without needing any abuse-of-form argument. The rubric judge catches this mismatch ($s^{(A)}\!=\!9$ vs $5$), yielding $\Delta_A\!=\!4 > \Delta_B\!=\!2$. \textbf{Signal: in law, the correct doctrinal pathway matters as much as the correct conclusion.}}

\end{datacase}

\end{document}